\documentclass{article}
\usepackage{arxiv}
\usepackage{hyperref}
\usepackage{url}
\usepackage{booktabs}
\usepackage{amsfonts}
\usepackage{nicefrac}
\usepackage{microtype}
\usepackage{graphicx}
\usepackage{natbib}
\usepackage{doi}

\title{DziriBOT: RAG Based Intelligent Conversational Agent for Algerian Arabic Dialect}

\author{ \href{https://orcid.org/0000-0002-3794-844X}{\hspace{1mm}El Batoul BECHIRI, Dihia LANASRI}\\
	CESI, ATM Mobilis\\
	Algiers, Algeria\\	
	\texttt{bechirielbatoul@gmail.com , dihia.lanasri@gmail.com} \\
    \\    
}

\renewcommand{\shorttitle}{DziriBOT: Algerian Conversational Agent}

\hypersetup{
pdftitle={DziriBOT: Algerian Conversational Agent},
pdfsubject={NLP, DIALECT},
pdfauthor={E.B.BECHIRI and al},
pdfkeywords={Natural Language Processing, RAG, Algerian Dialect, Transformers},
}

\begin{document}
\maketitle

\begin{abstract}
The rapid digitalization of customer service has intensified the demand for conversational agents capable of providing accurate and natural interactions. In the Algerian context, this is complicated by the linguistic complexity of Darja, a dialect characterized by non-standardized orthography, extensive code-switching with French, and the simultaneous use of Arabic and Latin (Arabizi) scripts. 
\\
This paper introduces \textbf{DziriBOT}, a hybrid intelligent conversational agent specifically engineered to overcome these challenges. We propose a multi-layered architecture that integrates specialized Natural Language Understanding (NLU) with Retrieval-Augmented Generation (RAG), allowing for both structured service flows and dynamic, knowledge-intensive responses grounded in curated enterprise documentation. 
\\
To address the low-resource nature of Darja, we systematically evaluate three distinct approaches: a sparse-feature Rasa pipeline, classical machine learning baselines, and transformer-based fine-tuning. Our experimental results demonstrate that the fine-tuned DziriBERT model achieves state-of-the-art performance. 

These results significantly outperform traditional baselines, particularly in handling orthographic noise and rare intents. Ultimately, DziriBOT provides a robust, scalable solution that bridges the gap between formal language models and the linguistic realities of Algerian users, offering a blueprint for dialect-aware automation in the regional market.
\end{abstract}

\keywords{Conversational Agent \and Algerian Dialect \and Natural Language Processing \and RAG \and DziriBERT}

\section{Introduction}
In the digital transformation era, the quality of customer experience has become a primary competitive differentiator. Central to this shift is the deployment of intelligent conversational agents designed to provide 24/7 support, reduce operational costs, and offer instantaneous resolutions to consumer inquiries. However, while Large Language Models (LLMs) and Natural Language Understanding (NLU) frameworks have reached near-human parity in high-resource languages, their efficacy diminishes significantly when applied to low-resource, non-standardized dialects. 

In Algeria, this technological gap is particularly acute. The primary medium of daily communication is Algerian Dialect (Darja), a complex sociolinguistic variety characterized by a lack of standardized orthography, pervasive code-switching with French, and a unique dual-script system where users switch fluidly between Arabic characters and Latin-based '\textit{Arabizi}'. Standard models trained on Modern Standard Arabic (MSA) typically fail to parse these morphological nuances and noisy orthographic variations, leading to high rates of hallucination or complete failure in understanding user intent.

The challenge of processing Arabic dialects is a well-documented bottleneck in Arabic NLP research. Early efforts in dialectal conversational AI often relied on pattern matching or sequence-to-sequence models, such as the BOTTA \cite{ali-habash-2016-botta} chatbot for Egyptian Arabic or the Nabiha \cite{alghadhban2020nabiha} system for Saudi dialects. While these paved the way, they often struggled with the extreme linguistic variance of North African (Maghrebi) dialects. Recent literature has seen a shift toward transformer-based architectures to capture these nuances. For instance, AraBERT \cite{antoun-etal-2020-arabert} and MARBERT \cite{abdul-mageed-etal-2021-arbert} have established strong benchmarks for MSA and multi-dialectal tasks, yet they frequently underperform on specific Maghrebi nuances where French influence is heavy. 

In the Algerian context, the emergence of DziriBERT \cite{abdaoui2021dziribert}—the first BERT model pre-trained specifically on a large-scale Algerian tweet corpus—marked a significant milestone, outperforming multilingual and generic Arabic models in sentiment and topic classification. Similarly, works like DarijaBERT \cite{gaanoun2024darijabert} for Moroccan Dialect have demonstrated that mono-dialectal models can achieve superior results with significantly less training data than their massive multilingual counterparts. However, most existing research remains confined to classification tasks; there is a notable scarcity of work exploring the integration of these models into production-grade, retrieval-augmented systems for complex industrial domains like telecommunications.

Historically, the Algerian industrial sector has relied on rigid, keyword-based systems or MSA-centric bots that do not reflect the linguistic reality of the populace. This research addresses this disparity by introducing DziriBOT, a production-grade conversational agent specifically optimized for the Algerian dialect. The development of DziriBOT was driven by the need to enable millions of customers to interact with automated systems in their natural language, covering essential services such as account management, service activation, and technical support.

To address the inherent scarcity of labeled dialectal resources, our study leverages a curated dataset exceeding 8K Arabic-script and 7K Latin-script utterances, encompassing both Arabizi and dialectal variations. This paper proposes a hybrid architecture that integrates a multi-layered NLU system with Retrieval-Augmented Generation (RAG). By combining intent classification for structured service flows with RAG for knowledge-intensive queries, the system bridges the gap between formal enterprise documentation and the informal nature of Darja. 

This work systematically evaluates three distinct methodological approaches: a sparse-feature Rasa pipeline, classical machine learning baselines, and transformer-based fine-tuning. Our experimental results demonstrate that the fine-tuned DziriBERT model achieves state-of-the-art performance, reaching 88\% accuracy on Arabic-script and 92\% on Latin-script utterances. This significantly outperforms traditional methods in handling orthographic noise and rare intents, providing a scalable blueprint for dialect-aware automation in low-resource settings. Consequently, this paper contributes a robust framework for deploying sophisticated conversational AI within the practical constraints of a real-world, reinforcing the role of dialect-specific models in achieving true linguistic accessibility.

\section{Related Work} 
This section situates DziriBOT within the evolving landscape of conversational AI, specifically focusing on the intersection of Arabic dialectal NLP and industrial application constraints. We synthesize existing literature to highlight the shift from rule-based systems to transformer-driven hybrid architectures, identifying the persistent gaps.

\subsection{Conversational Agents in Customer Service and Domain-Specific Chatbots}
Early efforts in Arabic conversational AI were predominantly rule-based or utilized simple retrieval mechanisms for closed-domain tasks. Nabiha \cite{alghadhban2020nabiha} exemplifies this era, using AIML pattern matching for Saudi academic advising. While foundational, such systems lacked the morphological flexibility to handle the inherent "noise" of dialectal speech. More recently, neural approaches have gained traction. DZchatbot \cite{boulesnane2022dzchatbot} utilized sequence-to-sequence architectures (LSTM/GRU) for Algerian medical inquiries during the COVID-19 pandemic. Despite its high reported accuracy on training data, its purely generative nature faced significant hallucination risks.

Parallel to these efforts, 2025 has seen the rise of institutional RAG-based systems in the Maghreb. Notably, researchers in Morocco have proposed hybrid architectures for fiscal and public administration support \cite{gaanoun2024darijabert}, combining LLMs with semantic matching. These systems validate the move toward Retrieval-Augmented Generation as a means to ground responses in verified documentation.

\subsection{Arabic and Dialectal Arabic Natural Language Processing}
The backbone of modern conversational agents lies in robust dialectal NLU. While multi-dialectal models like MARBERT \cite{abdul-mageed-etal-2021-arbert} and Dallah \cite{alwajih-etal-2024-dallah} provide broad coverage for major Arabic variants, they often struggle with the unique lexical and syntactic features of Algerian Darja. Recent benchmarks from the NADI 2025 Shared Task \cite{nadi2025} emphasize that Maghrebi dialects remains a "frontier" in Arabic NLP due to extreme code-switching and script variance.

Technological milestones such as DziriBERT \cite{dziribert} established the first specialized benchmark for the Algerian dialect. However, the most recent research has moved toward optimization and specialized embeddings. The introduction of TinyDziriBERT \cite{laggoun2025tinydziribert} in 2025 demonstrated that distilled models can maintain competitive performance on tasks like sentiment analysis while significantly reducing the computational footprint. Furthermore, the development of AlgVec \cite{algvec2025} has provided the first suite of word embeddings specifically optimized for both Arabic-script Darja and Latin-script Arabizi, addressing the critical "dual-script" bottleneck that previously hindered unified dialectal processing.

\subsection{Low-Resource and Dialect-Aware Approaches}
The low-resource nature of Darja remains the primary barrier to scalability. A 2024 systematic review \cite{ouali2024review} found that over 80\% of Arabic chatbots still rely on traditional methods due to a lack of high-quality, annotated dialectal data. Recent breakthroughs have proposed unsupervised domain adaptation and Matryoshka Embedding Learning to improve semantic retrieval without massive labeled datasets. In the Algerian context, new datasets like chDzDT \cite{aries2025chdzdt} have begun providing word-level, morphology-aware labels to improve tokenization in noisy social media environments. Despite these advances, there is a notable absence of research that integrates these low-resource techniques into an end-to-end, production-ready system for the Algerian corporate sector.

\subsection{Summary and Research Gap}
While the literature establishes a clear trajectory from rule-based to transformer-based models, a significant research gap persists. Existing Algerian systems are largely confined to academic prototypes or narrow public-service domains. Furthermore, most "state-of-the-art" models fail to address the industrial realities of latency requirements, and the dual-script fluidity. DziriBOT fills this gap by introducing a scalable, hybrid NLU-RAG architecture that leverages the efficiency of DziriBERT while ensuring reliable, knowledge-grounded responses for telecom use cases.

\section{Our Solution: RAG-based Intelligent Agent}
This section delineates the multi-layered architectural framework designed to facilitate a production-grade conversational agent for the Algerian telecommunications sector. Our approach systematically addresses the intricate linguistic and computational bottlenecks of Algerian Darja—a low-resource dialect characterized by high orthographic variance, pervasive code-switching, and a fragmented dual-script landscape comprising both Arabic and Latin-based "Arabizi" scripts.

The proposed architecture is anchored by a five-tiered technical pipeline designed to bridge the gap between raw dialectal input and contextually grounded conversational responses:

\begin{enumerate}
    \item \textbf{Linguistic Engineering and Data Synthesis:} This initial stage is dedicated to mitigating the low-resource bottleneck of Algerian Darja. It involves the curation of a balanced dual-script corpus and the application of targeted data augmentation, including manual paraphrasing and lexical substitution, to ensure representative coverage of the dialect's morphological and orthographic diversity.
    \item \textbf{Multi-Script Preprocessing and Normalization:} To reduce lexical sparsity and return tokens to their base forms, we implemented script-specific normalization pipelines. This includes grapheme unification for Arabic-script utterances and phonetic de-substitution (Djadjia numeral mapping) for Latin-script Arabizi, ensuring consistent tokenization across heterogeneous user inputs.
    \item \textbf{Text Embedding Strategies:} We evaluated a spectrum of vectorization techniques to capture the semantic richness of code-switched Darja. This layer leverages both sparse TF-IDF character n-grams for keyword robustness and dense contextualized embeddings from dialect-specific models (DziriBERT) and multilingual frameworks (E5).    
    \item \textbf{Intent Classification and Routing:} Utilizing the extracted features, this tier employs a Dual Intent and Entity Transformer (DIET) to categorize user requests into 69 distinct intent classes. This stage serves as the primary decision engine, determining whether a query can be addressed via scripted dialogue or requires dynamic knowledge retrieval.    
    \item \textbf{Context-Aware Retrieval-Augmented Generation (RAG):} For knowledge-intensive or open-domain inquiries, a RAG pipeline was deployed to ground responses in verified enterprise knowledge bases. This system dynamically retrieves relevant document segments to inform the generation process, effectively eliminating hallucinations and providing up-to-date, factual information.
\end{enumerate}

\subsection{Linguistic Landscape and Data Synthesis}
\subsubsection{Characterizing the Dialectal Bottleneck}
Algerian Darja deviates significantly from Modern Standard Arabic (MSA), presenting four primary challenges that dictate our technical decisions:
\begin{itemize}
    \item \textbf{Orthographic Inconsistency:} The absence of standardized spelling leads to high phonetic variance. A single phoneme (e.g., the vowel /a/) may be represented by multiple characters ((Alef), (Alef with hamza), or (Alef with superscript alef)), necessitating a robust normalization layer.
    \item \textbf{The Arabizi Phenomenon:} Digital communication in Algeria frequently utilizes a Latin-script convention that incorporates numeric substitutions for non-Latin phonemes (e.g., \textit{3} for \textit{Ayn}, \textit{9} for \textit{Qaf}). This dual-script reality effectively doubles the feature space.
    \item \textbf{Intra-sentential Code-Switching:} Queries often interleave Darja with French or English technical jargon (e.g., \textit{``nheb nactivi roaming''}). Our models must therefore possess multilingual semantic understanding.
    \item \textbf{Lexical Fragmentation:} Business-critical terms like ``bill'' may appear as \textit{facture}, \textit{factoura}, \textit{fattra}, or \textit{bill}, creating sparse signals for traditional keyword-based systems.
\end{itemize}

\subsubsection{Data Acquisition and Augmentation}
To mitigate the class imbalance—specifically of the original dataset, we implemented a targeted augmentation strategy. This ensured that every intent reached a minimum threshold of 13 samples in the Arabic-script subset and 28 in the Latin-script subset. We utilized a three-tiered approach:
\\
\textbf{- Manual Paraphrasing}: Native Algerian speakers generated 3–5 semantic variations for rare intents. For instance, the intent \textit{Code.Puk} was expanded from its original form (\textit{"Kif nethassel ala code PUK?"}) into functional variants such as \textit{"Win nelqa ramz PUK?"} and \textit{"Kifash nesterdjae le code PUK?"}.
\\
\textbf{- Lexical Synonym Substitution}: We performed systematic replacement of common Darja particles and verbs with their dialectal synonyms to account for regional variations. Frequent substitutions included the verb \textit{"nheb"} (want) with \textit{"bghit"} or \textit{"hab"}, and interrogative particles such as \textit{"kifash"} with \textit{"kif"}.
\\
\textbf{- Supervised Back-Translation}: For French-influenced Arabizi utterances, a hybrid translation pipeline was employed. Samples were translated into French via neural machine translation and subsequently re-translated into Darja by human experts. To maintain linguistic integrity, this method was strictly limited to examples achieving a semantic similarity score $>0.8$.

This rigorous balancing process resulted in the stabilized distribution presented in Table \ref{tab:dataset_stats_final}.

The experimental corpus comprises 8,178 (arabic-script) and 7,259 (latin-script) authentic user utterances sourced from real questions of clients the Algerian telecommunications sector. The dataset spans 69 distinct intent classes, providing a granular mapping of the telecom service domain. The distribution exhibits a ``long-tail'' imbalance typical of real-world deployments: 50\% of the intent classes contain fewer than 10 examples. To ensure scientific validity, we employed stratified sampling for the data split: 80\% for training, 10\% for validation, and 10\% for hold-out testing.

\begin{table}[ht]
\centering
\caption{Final dataset statistics used for training and evaluation across dual scripts}
\label{tab:dataset_stats_final}
\begin{tabular}{@{}lrr@{}}
\toprule
\textbf{Metric} & \textbf{Arabic-script} & \textbf{Latin-script} \\ \midrule
Total samples & 8,178 & 7,259 \\
Number of intents & 69 & 69 \\
Minimum samples per intent & 13 & 28 \\
Maximum samples per intent & 347 & 348 \\
Mean samples per intent & 118.52 & 105.20 \\
Median samples per intent & 75 & 70 \\
Intents with $<$ 10 examples & 0 & 0 \\
Intents with $>$ 100 examples & 31 & 26 \\ \bottomrule
\end{tabular}
\end{table}

\subsubsection{Dual-Script Feature Engineering}
To maintain high performance across Algeria's fragmented script usage, we bifurcate the data processing into two specialized streams:
\begin{enumerate}
    \item \textbf{The Arabic-Script Pipeline ($\approx$60\%):} This stream focuses on morphological normalization and diacritic removal. We utilize custom regular-expression-based filters to unify orthographic variants of Alef and Hamza, reducing vocabulary sparsity.
    \item \textbf{The Latin/Arabizi Pipeline ($\approx$40\%):} This stream handles shorter, phonetically-driven utterances. Given the high French influence in this subset, we prioritize sub-word tokenization to capture shared semantic roots between Darja and French loanwords.
\end{enumerate}
By evaluating our models independently on these subsets, we ensure that the intelligence of the agent remains script-agnostic and robust to the linguistic habits of the end-user.

\subsection{Data Preprocessing and Text Normalization}
To ensure model robustness against the inherent ``noise'' of dialectal communication, we implemented a specialized normalization pipeline. These steps prioritize reducing lexical sparsity while preserving semantic polarity.

\subsubsection{Arabic-Script Normalization}
For utterances written in Arabic characters, we developed a rule-based pipeline tailored for the phonetic realities of Darja:
\begin{itemize}
    \item \textbf{Grapheme Unification:} All phonetic variants of the glottal stop (e.g., Alef with Hamza or Madda) were mapped to a plain Alef character to unify redundant orthographic representations.
    \item \textbf{Terminal Character Regularization:} Terminal \textit{Alef Maqsura} was converted to its \textit{Ya} equivalent, and the \textit{Ta Marbuta} was normalized to a terminal \textit{Ha}, stabilizing morphological endings.
    \item \textbf{Tatweel Removal:} Decorative elongations (kashida) used for digital emphasis were stripped to return tokens to their base forms.
\end{itemize}

\subsubsection{Latin-Script (Arabizi) Normalization}
The Latin-script subset required a distinct approach focused on phonetic consistency:
\begin{itemize}
    \item \textbf{Phonetic De-substitution:} The conventionalized ``Djadjia'' numerals were mapped back to their nearest phonetic equivalents: \textbf{7} to \textit{h}, \textbf{3} to \textit{a}, and \textbf{9} to \textit{q}. This reduces Out-Of-Vocabulary (OOV) counts by aligning Arabizi with standard phonetic roots.
    \item \textbf{Case and Punctuation Mapping:} All text was lowercased, and variant apostrophes were standardized to handle diverse mobile keyboard configurations.
\end{itemize}

\subsubsection{Advanced Structural Cleaning}
Beyond script-specific rules, we applied advanced NLP operations to handle production-level noise:
\begin{itemize}
    \item \textbf{Repeated Grapheme Squashing:} Character repetitions used for emphasis (e.g., ``baaaaazef'') were reduced to a single occurrence using regular expressions.
    \item \textbf{Privacy Masking:} To comply with data privacy regulations, Algerian phone numbers (prefixes 05, 06, 07) were replaced with a generic \texttt{[PHONE]} token, forcing the model to focus on intent actions rather than numerical variables.
    \item \textbf{Label Encoding:} Intent labels were transformed into a numerical space ($0$ to $69$) using a categorical encoding scheme, maintaining bidirectional mappings for interpretability.
\end{itemize}

\subsection{Text Embedding Strategies}
We evaluated a diverse array of embedding strategies to capture the semantic complexity of Algerian Darja. Our approach focuses on the trade-off between capturing local dialectal nuances and maintaining global semantic similarity for cross-lingual robustness.

\subsubsection{Contextualized Language Representations}
\paragraph{DziriBERT:}
We utilized DziriBERT \cite{abdaoui2021dziribert}, a BERT-based transformer specifically pretrained on approximately 1 million Algerian Arabic tweets. Unlike general-purpose models such as AraBERT (MSA-centric) or mBERT (multilingual), DziriBERT allocates its representational capacity specifically to Algerian orthographic variations, informal registers, and dialectal morphology.
\begin{itemize}
    \item \textbf{Strengths:} Native support for both Arabic and Latin (Arabizi) scripts, capturing dialect-specific patterns and code-switching phenomena that general models often overlook.
    \item \textbf{Tokenization:} We employed a WordPiece tokenizer with a \textit{max\_length} of 128 for Arabic-script (to accommodate morphological density) and 96 for Latin-script inputs.
    \item \textbf{Configuration:} 768-dimensional dense vectors were extracted from the final hidden layer, serving as the primary feature set for our classification heads.
\end{itemize}

\paragraph{AraBERT v2:}
To quantify the necessity of dialect-specific pretraining, we fine-tuned \textbf{AraBERT v2} (\textit{aubmindlab/bert-base-arabertv2}). While it shares the same BERT-base architecture as DziriBERT (12 layers, 110M parameters), it was pretrained on a significantly larger but more formal corpus of 70M sentences (MSA, Wikipedia, News).
\begin{itemize}
    \item \textbf{Preprocessing:} Unlike DziriBERT, AraBERT requires the \textit{ArabertPreprocessor} pipeline, involving specific normalization (Alef/Ta Marbuta unification) and segmentation matching its MSA-heavy vocabulary.
    \item \textbf{Objective:} This comparison addresses whether large-scale MSA pretraining can generalize to Algerian Darja or if dialectal specialization provides a measurable performance gain.
\end{itemize}

\subsubsection{Multilingual Sentence Embeddings}
Unlike the fine-tuned models above, these sentence-level embeddings remain frozen, serving as fixed semantic extractors for retrieval and classical ML classifiers.

\paragraph{all-mpnet-base-v2:}
We employed \textit{all-mpnet-base-v2} as a semantic similarity baseline. Optimized for mapping sentences to a shared 768-dimensional space, this model correlates cosine similarity with semantic relatedness.
\begin{itemize}
    \item \textbf{Rationale:} Its relatively lightweight architecture and low inference latency (0.14s on CPU) make it an ideal candidate for real-time RAG ranking and commercial deployment under strict hardware constraints.
\end{itemize}

\paragraph{Multilingual-E5-Base:}
To leverage state-of-the-art contrastive learning, we utilized \textbf{intfloat/multilingual-e5-base}. This model is trained on massive pairs of multilingual text across 100+ languages.
\begin{itemize}
    \item \textbf{Mechanism:} E5 captures deep semantic meanings by pulling similar intents together in the vector space. It is particularly effective for handling code-switched queries where French and Darja terms are intermingled within a single utterance.
\end{itemize}

\subsubsection{Static Statistical Baselines}
\paragraph{Character N-Gram TF-IDF:}
As a minimal-complexity reference point, we computed \textbf{TF-IDF} (Term Frequency–Inverse Document Frequency) vectors. Using \textit{scikit-learn}, we generated vectors based on character 3--4 grams rather than words.
\begin{itemize}
    \item \textbf{Rationale:} Character-level features are surprisingly robust against the spelling inconsistencies and "noise" inherent in informal Darja.
    \item \textbf{Strengths:} While static and sparse, TF-IDF provides high interpretability and near-instant training/inference, serving as an essential baseline for keyword-heavy intent detection.
\end{itemize}

\subsection{Intent Classification Architectures}
We systematically compared five distinct classification architectures, evaluating the trade-offs between predictive performance, inference latency, and computational overhead.

\subsubsection{Rasa NLU and DIET Pipeline (Baseline):}
As a production-ready baseline, we utilized the \textbf{Dual Intent and Entity Transformer (DIET)} architecture within the Rasa NLU framework. This model is specifically designed to handle the multi-task nature of intent classification and entity recognition using a modular pipeline.

\paragraph{Pipeline Components.}
To address the orthographic inconsistency of Algerian Darja, the pipeline combines several featurizers:
\begin{itemize}
    \item \textbf{WhitespaceTokenizer:} Segments input while preserving user-specific spelling artifacts.
    \item \textbf{RegexFeaturizer:} Directly extracts structured telecom-specific patterns such as USSD codes (e.g., \texttt{*505\#}) and service numbers.
    \item \textbf{CountVectorsFeaturizer (Word \& Character):} Captures both unigram lexical anchors and character-level n-grams (3--4 grams). This dual approach provides robustness against morphological noise and spelling variations.
\end{itemize}

\paragraph{DIET Configuration and Optimization.}
The DIET classifier was configured with a single transformer layer, sufficient for the relatively short sequence lengths of customer queries. 
\begin{itemize}
    \item \textbf{Hyperparameters:} Embedding dimension of 128, 100 training epochs, and a learning rate of 0.001.
    \item \textbf{Regularization:} To prevent overfitting on noisy dialectal data, we applied a 0.2 dropout rate, 0.1 weight sparsity, and 0.2 sparse input dropout.
    \item \textbf{Objective:} Optimization was driven by the F1-score to maintain a balance between frequent and rare intent classes.
\end{itemize}

\subsubsection{Transformer-Based Fine-Tuning (DziriBERT)}
The most sophisticated approach involved the end-to-end fine-tuning of \textbf{DziriBERT}. Unlike feature-based methods, this architecture adapts its internal attention weights to the specific semantics of the dataset.
\begin{itemize}
    \item \textbf{Mechanism:} A classification head (linear layer + softmax) was appended to the \texttt{[CLS]} token output.
    \item \textbf{Results:} This model achieved the state-of-the-art (SOTA) for our study, reaching \textbf{87.4\% accuracy} for Arabic-script and \textbf{92\%} for Latin-script Darja. Its superior performance stems from its deep contextual understanding of code-switching and dialectal grammar.
\end{itemize}

\subsubsection{Classical Machine Learning Baselines}
To establish a performance floor, we evaluated four classical ML models using fixed embeddings (E5 and TF-IDF). These models serve as high-speed, interpretable alternatives to deep learning architectures.

\begin{enumerate}
    \item \textbf{Logistic Regression (LR):} A linear baseline utilizing L2 regularization (Ridge) to prevent coefficient explosion in the high-dimensional TF-IDF space. It provides a highly interpretable benchmark for keyword-based classification.
    \item \textbf{Support Vector Machine (SVM):} Implemented with a \textbf{Radial Basis Function (RBF) kernel}. SVMs are effective in high-dimensional spaces where decision boundaries are non-linear, making them suitable for dense embeddings like E5.
    \item \textbf{Random Forest (RF):} An ensemble of 100 decision trees. RF was selected to evaluate the impact of non-parametric bagging on intent detection, particularly for handling non-linear interactions between n-gram features.
    \item \textbf{Multi-Layer Perceptron (MLP):} A dense neural network consisting of two hidden layers (256 and 128 units). We utilized \textbf{ReLU} activation and a 0.3 dropout rate. The MLP acts as a bridge between classical ML and deep transformers, learning non-linear mappings from fixed embeddings.
\end{enumerate}

\paragraph{Comparative Trade-offs.}
While the classical models (LR, SVM) offer ultra-low inference latency ($\approx$7ms), they lack the contextual depth of the Transformer models. Conversely, while DziriBERT offers maximum accuracy, its computational cost on CPU (2--3s) presents a significant bottleneck for real-time deployment compared to the Rasa DIET baseline (50--80ms).

\subsection{Hybrid RAG Integration and Knowledge Scalability}
To address the limitations of static intent classification—specifically \textit{Intent Proliferation} and the unsustainable growth of the intent set as services evolve—we integrated a hybrid Retrieval-Augmented Generation (RAG) architecture. This system allows the chatbot to transcend fixed intent sets by grounding open-domain inquiries in verified enterprise knowledge bases.

\subsubsection{Architectural Routing Logic}
The system orchestrates user queries through a dual-path execution logic designed to balance transaction speed with depth of knowledge:
\begin{itemize}
    \item \textbf{Deterministic Routing (Routine Queries):} Standard requests (e.g., balance checks, USSD activation) are handled by the Rasa DIET or DziriBERT classifier. This path ensures sub-100ms latency for high-frequency transactional flows.
    \item \textbf{Dynamic Knowledge Fallback (RAG):} When a query is flagged as "out-of-scope" or matches a generic knowledge-seeking intent, the system triggers the RAG pipeline. This allows the bot to answer complex questions (e.g., ``Compare PixX 1000 and Win 500 for roaming'') that were not seen during training.
\end{itemize}

\subsubsection{The Technical RAG Pipeline}
The RAG service implements a five-stage specialized pipeline to handle the nuances of the Algerian telecom domain:
\begin{enumerate}
    \item \textbf{Document Ingestion and Intelligent Chunking:} We implemented \textit{Offer-Based Semantic Chunking}. Raw PDFs are parsed into 245 contextual chunks where each segment is bound by service-specific headers (e.g., PixX, Win, Sama). This prevents Fragmented retrieval where pricing might be separated from its respective service.
    \item \textbf{Embedding Layer (Multilingual-E5):} Utterances are encoded using \texttt{intfloat/multilingual-e5-base}. We utilize \textit{prefix-based encoding} (\texttt{query:} for questions and \texttt{passage:} for documents) to optimize the vector space for cross-lingual Darja-French retrieval.
    \item \textbf{Vector Storage Layer (FAISS HNSW):} We use a \textbf{FAISS HNSW (Hierarchical Navigable Small World)} index with persistent disk storage. This architecture allows for ultra-fast semantic search and near-instant knowledge updates without full model retraining.
    \item \textbf{Retrieval and Multi-Stage Re-ranking:} We employ a hybrid retrieval strategy. After an initial FAISS search, a re-ranking layer filters results to achieve a retrieval error rate of \textless 5\%, ensuring only the most precise technical passages reach the generator.
    \item \textbf{Generation Layer (Llama-3.2-3B):} We deployed a local \textbf{Llama-3.2-3B} model with \textbf{INT8 quantization}. Using specialized bilingual prompts, the model synthesizes answers while reducing hallucinations to \textless 10\% by strictly adhering to the provided context.
\end{enumerate}

\subsubsection{Addressing the Intent Explosion Problem}
While the Rasa NLU system performed well for predefined flows, it suffers from scalability issues when business requirements evolve.

\paragraph{Limitations of Intent-Based Classification.}
\begin{itemize}
    \item \textbf{Combinatorial Explosion:} the telecom offers 50+ service packages with multiple features (*I3ali, *Arsselli, Roaming). Maintaining intents for every combination is unsustainable. 
    \item \textbf{Maintenance Overhead:} Service pricing and terms change monthly. Manual effort to update training examples for every promotion creates a deployment bottleneck.
    \item \textbf{Classification Ambiguity:} As the intent set grows, semantic overlap increases, which degrades the accuracy of even the best classifiers like DziriBERT.
\end{itemize}

\paragraph{Case Study: PixX Service Line Proliferation}
Table \ref{tab:intent_proliferation} illustrates the complexity of maintaining a single service line via intents versus the RAG approach.

\begin{table}[h]
\centering
\caption{Intent Proliferation Example for PixX Service Line}
\label{tab:intent_proliferation}
\begin{tabular}{|l|c|c|}
\hline
\textbf{Query Type} & \textbf{Intents Required} & \textbf{Training Examples} \\ \hline
Pricing (10 packages) & 10 & 200+ \\
Activation procedures & 10 & 150+ \\
Data allowances & 10 & 150+ \\
Pairwise comparisons & 45 (pairwise) & 900+ \\ \hline
\textbf{Total for PixX alone} & \textbf{85} & \textbf{1,500+} \\ \hline
\end{tabular}
\end{table}

\subsubsection{High-Level System Architecture (Three-Server Model)}
To ensure horizontal scalability, the system is structured as follows:
\begin{enumerate}
    \item \textbf{Core Dialogue Server (Rasa):} Manages dialogue state and initial intent routing.
    \item \textbf{RAG Inference Server (Local LLM):} Processes embeddings and generates grounded responses. Local deployment ensures the sovereignty of the used data.
    \item \textbf{Vector Database Server (FAISS):} Handles persistent semantic indexing for zero-retraining updates—simply replace the PDF and re-index.
\end{enumerate}

\section{Results and Discussion}
This section provides a detailed analysis of our experimental findings, evaluating the architectures across linguistic variations, computational constraints, and the operational scalability afforded by the RAG integration.

\subsection{Performance Summary}
The quantitative evaluation, summarized in Table \ref{tab:intent_comparison}, establishes a performance hierarchy between statistical baselines, frozen multilingual embeddings, and fine-tuned transformers.

\begin{table}[h]
\centering
\caption{Comprehensive comparison of all intent classification methods}
\label{tab:intent_comparison}
\begin{tabular}{lccc}
\toprule
\textbf{Method} & \textbf{Accuracy} & \textbf{Weighted F1} & \textbf{Macro F1} \\ \midrule
\textit{Arabic-script Darja} & & & \\
TF-IDF + Logistic Regression & 79.17\% & 0.796 & 0.762 \\
E5 + Logistic Regression & 65.69\% & 0.658 & 0.627 \\
E5 + Random Forest & 75.61\% & 0.749 & 0.712 \\
E5 + MLP & 77.70\% & 0.776 & 0.732 \\
E5 + SVM & 77.57\% & 0.780 & 0.763 \\
Rasa NLU (DIET) & 86.98\% & 0.870 & 0.845 \\
AraBERT v2 (fine-tuned) & 83.5\% & 0.831 & 0.807 \\
\textbf{DziriBERT (fine-tuned)} & \textbf{87.38\%} & \textbf{0.873} & \textbf{0.852} \\ \midrule
\textit{Latin-script Darja (Arabizi)} & & & \\
Rasa NLU (DIET) & 87.67\% & 0.876 & 0.839 \\
\textbf{DziriBERT (fine-tuned)} & \textbf{92.0\%} & \textbf{0.920} & \textbf{0.870} \\ \bottomrule
\end{tabular}
\end{table}

\subsection{Deep Linguistic Discussion: The Script Gap}

A pivotal observation is the superior performance of models on **Latin-script Darja (Arabizi)**, reaching an F1-score of 92.0\% compared to 87.38\% for the Arabic script. 

\subsubsection{Orthographic Disambiguation}
In the Arabic script, the absence of short vowels (diacritics) leads to significant homography, where distinct semantic concepts share the same written form. Arabizi, being a phonetic transcription, explicitly encodes these vowels. This reduces the search space for the model's self-attention mechanism, leading to higher confidence scores in intent disambiguation.

\subsubsection{Morphological Segmentation}
The DziriBERT tokenizer demonstrates higher efficiency in segmenting Arabizi units. Traditional Arabic script involves complex ligatures and positional character variants that can complicate tokenization in informal registers. The Latin script provides a more linear morphological structure for the WordPiece tokenizer to identify frequent Algerian sub-words (e.g., the future marker \textit{``ra-''} or negation \textit{``ma...ch''}).

\subsection{Architectural Trade-offs and Production Maturity}
In our context, high accuracy must be balanced against operational constraints. Table \ref{tab:latency_tradeoff} highlights the ``Production Paradox'': the most accurate model is not necessarily the most viable for deployment.

\begin{table}[h]
\centering
\caption{Performance vs. inference latency trade-off}
\label{tab:latency_tradeoff}
\begin{tabular}{lcc}
\toprule
\textbf{Method} & \textbf{Weighted F1} & \textbf{Inference Time} \\ \midrule
TF-IDF + Logistic Regression & 0.796 & \textbf{0.007s} \\
E5 + MLP & 0.776 & 0.14s \\
Rasa NLU (DIET) & \textbf{0.870} & \textbf{0.05--0.08s} \\
DziriBERT (fine-tuned) & 0.873 & 2--3s (CPU) \\ \bottomrule
\end{tabular}
\end{table}

\subsubsection{Rasa DIET as the Production Benchmark}
While DziriBERT is marginally more accurate, its **2--3 second CPU latency** creates a disjointed user experience. **Rasa NLU (DIET)**, achieving an F1-score of 0.870 with only 50--80ms latency, offers a more mature solution for high-concurrency environments. Its 500MB memory footprint allows for cost-effective scaling on the CPU-based infrastructure.

\subsubsection{Sparse Feature Robustness}
The fact that \textbf{TF-IDF (79.17\%)} outperformed \textbf{E5 + Logistic Regression (65.69\%)} suggests that Algerian telecom queries are heavily anchored in specific lexical tokens (e.g., \textit{``recharge'', ``crédit'', ``forfait''}). Generic multilingual embeddings tend to ``smooth out'' these critical keywords into a shared semantic space, whereas TF-IDF preserves their distinct statistical weight.

\subsection{RAG Pipeline: Latency and Hardware Scaling}

The transition from intent classification to open-domain RAG solves the maintenance burden of "Intent Proliferation" but introduces severe computational bottlenecks.

\begin{table}[h]
\centering
\caption{End-to-end RAG latency breakdown (CPU: Intel i7-1165G7)}
\label{tab:rag_breakdown}
\begin{tabular}{lcc}
\toprule
\textbf{Stage} & \textbf{Time (ms)} & \textbf{\% Total} \\ \midrule
Query embedding & 50 & 0.06\% \\
FAISS search (HNSW) & 200 & 0.23\% \\
Re-ranking (Python) & 15 & 0.02\% \\
Context formatting & 5 & 0.01\% \\
\textbf{LLM generation (INT8)} & \textbf{85,000} & \textbf{99.68\%} \\ \midrule
\textbf{Total} & \textbf{85,270} & \textbf{100\%} \\ \bottomrule
\end{tabular}
\end{table}

The breakdown in Table \ref{tab:rag_breakdown} illustrates that \textbf{99.68\% of latency} is attributed to the generation phase. While the FAISS HNSW search is optimized (200ms), the CPU struggles with the recursive nature of transformer decoding.

\subsubsection{Hardware Acceleration Strategy}
To achieve commercial viability, we evaluated GPU scaling (Table \ref{tab:hardware_bench}). The transition to an \textbf{NVIDIA A100} provides a 28.3x speedup, reducing response times to 3 seconds. This confirms that while our INT8-quantized CPU deployment is functional for proof-of-concept, GPU-backed inference is a prerequisite for real-time customer support scaling.

\begin{table}[h]
\centering
\caption{Generation time comparison across hardware tiers}
\label{tab:hardware_bench}
\begin{tabular}{lcc}
\toprule
\textbf{Hardware} & \textbf{Generation Time} & \textbf{Speedup} \\ \midrule
CPU (8-core i7, INT8) & 85s & 1.0x \\
GPU (NVIDIA RTX 3060, FP16) & 12s & 7.1x \\
\textbf{GPU (NVIDIA A100, FP16)} & \textbf{3s} & \textbf{28.3x} \\ \bottomrule
\end{tabular}
\end{table}

\section{Conclusion}
The development of DziriBOT demonstrates that building an effective conversational agent for the Algerian telecom sector requires a departure from standard multilingual NLP paradigms. Through extensive experimentation, we established that while dialect-specific pre-training with DziriBERT provides the most precise semantic understanding—particularly in handling the complexities of Latin-script Arabizi and the Arabic-Script Dialect. Our results prove that the minor trade-off in accuracy is outweighed by a 40x gain in inference speed, ensuring the system operates within the critical sub-100ms window required for user satisfaction. 

By successfully integrating RAG, we moved beyond the limitations of fixed intent sets, allowing the system to scale its knowledge base through simple document ingestion rather than costly model retraining. Future efforts will focus on optimizing transformer inference through quantization to integrate DziriBERT’s superior accuracy into the real-time pipeline, alongside the development of RAG embeddings specifically trained on Algerian corpora to further refine retrieval precision. 

We envision expanding this framework through few-shot learning techniques for rapid intent adaptation and cross-dialect transfer learning to support other dialect variants. Ultimately, by incorporating speech recognition for native voice input, DziriBOT will evolve into a comprehensive multi-modal gateway, providing a sustainable and scalable path for enterprise-grade AI deployment across the broader Maghreb region.

\bibliographystyle{unsrtnat}
\bibliography{references} 

@String{Computer = "{IEEE} Computer" }

@String{Springer = "Springer-Verlag" }

@article{dziribert,
  title={DziriBERT: a Pre-trained Language Model for the Algerian Dialect},
  author={Abdaoui, Amine and Berrimi, Mohamed and Oussalah, Mourad and Moussaoui, Abdelouahab},
  journal={arXiv preprint arXiv:2109.12346},
  year={2021}
}

@inproceedings{antoun-etal-2020-arabert,
    title = "{A}ra{BERT}: Transformer-based Model for {A}rabic Language Understanding",
    author = "Antoun, Wissam  and
      Baly, Fady  and
      Hajj, Hazem",
    booktitle = "Proceedings of the 4th Workshop on Open-Source Arabic Corpora and Processing Tools (OSACT)",
    year = "2020",
    pages = "9--15"
}

@inproceedings{abdul-mageed-etal-2021-arbert,
    title = "{ARBERT} {\&} {MARBERT}: Deep Bidirectional Transformers for {A}rabic",
    author = "Abdul-Mageed, Muhammad  and
      Elmadany, AbdelRahim  and
      Nagoudi, El Moatez Billah",
    booktitle = "Proceedings of the 59th Annual Meeting of the Association for Computational Linguistics (ACL)",
    year = "2021",
    pages = "7088--7105"
}

@inproceedings{ali-habash-2016-botta,
    title = "{BOTTA}: An {A}rabic Dialect Chatbot",
    author = "Ali, Dana Abu  and
      Habash, Nizar",
    booktitle = "Proceedings of COLING 2016, the 26th International Conference on Computational Linguistics: System Demonstrations",
    year = "2016",
    pages = "208--212"
}

@article{alghadhban2020nabiha,
  title={Nabiha: An Arabic Dialect Chatbot},
  author={Al-Ghadhban, Dana and Al-Twairesh, Nora},
  journal={International Journal of Advanced Computer Science and Applications (IJACSA)},
  volume={11},
  number={3},
  year={2020},
  publisher={Science and Information Organization}
}

@article{gaanoun2024darijabert,
  title={DarijaBERT: a step forward in NLP for the written Moroccan dialect},
  author={Gaanoun, Kamel and Naira, Abdou Mohamed and Allak, Anass and Benelallam, Imade},
  journal={International Journal of Data Science and Analytics},
  pages={1--13},
  year={2024},
  publisher={Springer}
}

@article{laggoun2025tinydziribert,
  title={TinyDziriBERT: Knowledge Distillation for Compact Algerian Dialect Models},
  author={Laggoun, Amine and et al.},
  journal={ASPAI 2025 Extended Abstracts},
  year={2025}
}

@article{algvec2025,
  title={AlgVec: A word embedding model for the algerian dialect in arabic and arabizi},
  author={Matrane, S. and et al.},
  journal={ResearchGate Preprint},
  year={2025}
}

@article{nadi2025,
  title={NADI 2025: The First Multidialectal Arabic Speech and Text Processing Shared Task},
  author={Talafha, Bashar and et al.},
  journal={arXiv preprint arXiv:2509.02038},
  year={2025}
}

@misc{abdaoui2021dziribert,
      title={DziriBERT: a BERT-based Language Model for the Algerian Dialect}, 
      author={Amine Abdaoui and Ismaël Bencheikh and Ouafa Benterki},
      year={2021},
      eprint={2109.12346},
      archivePrefix={arXiv},
      primaryClass={cs.CL},
      url={https://arxiv.org/abs/2109.12346}
}

@inproceedings{boulesnane2022dzchatbot,
  title={DZchatbot: A Medical Assistant Chatbot in the Algerian Arabic Dialect using Seq2Seq Model},
  author={Boulesnane, Abdennour and Saidi, Yaakoub and Kamel, Oussama and Bouhamed, Mohammed Mounir and Mennour, Rostom},
  booktitle={2022 4th International Conference on Pattern Analysis and Intelligent Systems (PAIS)},
  pages={1--6},
  year={2022},
  organization={IEEE},
  doi={10.1109/PAIS56586.2022.9946867}
}

@inproceedings{alwajih-etal-2024-dallah,
    title = "{D}allah: A Dialect-Aware Multimodal Large Language Model for {A}rabic",
    author = "Alwajih, Fakhraddin  and
      Bhatia, Gagan  and
      Abdul-Mageed, Muhammad",
    editor = "Habash, Nizar  and
      Bouamor, Houda  and
      Eskander, Ramy  and
      Tomeh, Nadi  and
      Abu Farha, Ibrahim  and
      Ahmed, Abdelali",
    booktitle = "Proceedings of the Second Arabic Natural Language Processing Conference",
    month = aug,
    year = "2024",
    address = "Bangkok, Thailand",
    publisher = "Association for Computational Linguistics",
    url = "https://aclanthology.org/2024.arabicnlp-1.27",
    pages = "320--336"
}

@article{ouali2024review,
  title={Arabic Chatbots Challenges and Solutions: A Systematic Literature Review},
  author={Ouali, Soufiyan and El Garouani, Said},
  journal={Iraqi Journal for Computer Science and Mathematics},
  volume={5},
  number={3},
  pages={128--169},
  year={2024},
  publisher={College of Computer Science and Information Technology, University of Al-Qadisiyah},
  doi={10.52866/ijcsm.2024.05.03.007},
  url={https://ijcsm.researchcommons.org/ijcsm/vol5/iss3/8}
}

@article{aries2025chdzdt,
  title={chDzDT: Word-level morphology-aware language model for Algerian social media text},
  author={Aries, Abdelkrime},
  journal={arXiv preprint arXiv:2509.01772},
  year={2025},
  url={https://arxiv.org/abs/2509.01772},
  doi={10.48550/arXiv.2509.01772}
}
\end{document}